\documentclass[11pt]{article}

\usepackage[final]{acl}

\usepackage{times}
\usepackage{latexsym}

\usepackage[T1]{fontenc}
\usepackage{xcolor}

\usepackage[utf8]{inputenc}

\usepackage{microtype}

\usepackage{inconsolata}

\usepackage{graphicx}

\usepackage{caption}
\usepackage{subcaption}
\usepackage{booktabs}
\usepackage[table]{xcolor}
\usepackage{hyperref}

\usepackage{amsmath}

\title{Towards Global Retrieval Augmented Generation: A Benchmark for Corpus-Level Reasoning}

\author{
Qi Luo$^{1,*}$, Xiaonan Li$^{1,*}$, Tingshuo Fan$^{1}$, Xinchi Chen$^{1,\dagger}$, Xipeng Qiu$^{1,2,\dagger}$\\
$^{1}$Fudan University \\
$^{2}$Shanghai Innovation Institutey \\
\texttt{qluo22@m.fudan.edu.cn}}

\begin{document}
\maketitle
\renewcommand{\thefootnote}{\fnsymbol{footnote}}
\footnotetext[1]{Equal contribution.}
\footnotetext[2]{Corresponding author.}

\begin{abstract}
Retrieval-augmented generation (RAG) has emerged as a leading approach to reducing hallucinations in large language models (LLMs). Current RAG evaluation benchmarks primarily focus on what we call local RAG: retrieving relevant chunks from a small subset of documents to answer queries that require only localized understanding within specific text chunks. However, many real-world applications require a fundamentally different capability—global RAG—which involves aggregating and analyzing information across entire document collections to derive corpus-level insights (e.g. ``What are the top 10 most cited papers in 2023?''). 
In this paper, we introduce GlobalQA—the first benchmark specifically designed to evaluate global RAG capabilities, covering four core task types: counting, extremum queries, sorting, and top-k extraction. Through systematic evaluation across different models and baselines, we find that existing RAG methods perform poorly on global tasks, with the strongest baseline achieving only 1.51 F1 score.
To address these challenges, we propose GlobalRAG, a multi-tool collaborative framework that preserves structural coherence through chunk-level retrieval, incorporates LLM-driven intelligent filters to eliminate noisy documents, and integrates aggregation modules for precise symbolic computation. On the Qwen2.5-14B model, GlobalRAG achieves 6.63 F1 compared to the strongest baseline's 1.51 F1, validating the effectiveness of our method. 
\end{abstract}

\section{Introduction}
\label{sec1}

\begin{figure}[t]
    \flushright
    \begin{subfigure}[b]{0.45\textwidth}
        \centering
        \includegraphics[width=\textwidth]{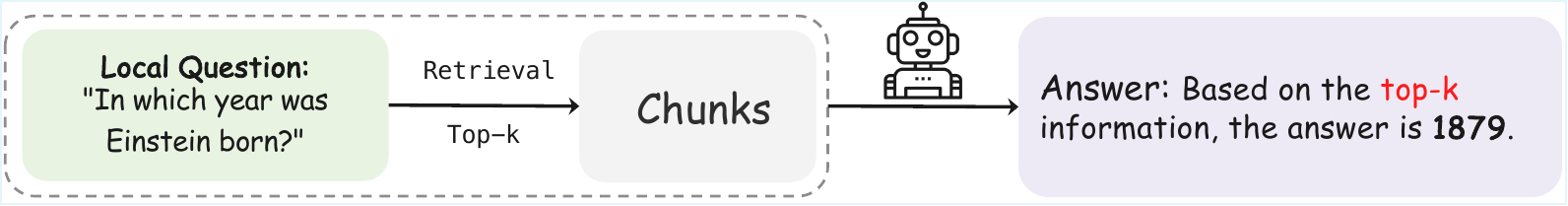}
        \caption{Local Query: Dense Retriever Works Well}
        \label{fig:local-query-rag}
    \end{subfigure}
    
    \vspace{0.02\textheight}
    
    \begin{subfigure}[b]{0.45\textwidth}
        \centering
        \includegraphics[width=\textwidth]{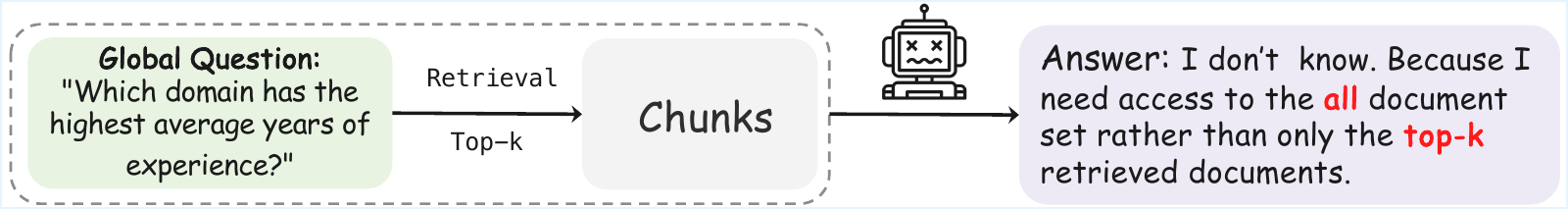}
        \caption{Global Query: Dense Retriever Fails}
        \label{fig:global-query-rag}
    \end{subfigure}
    \caption{Why Dense Retriever Fails on Global Queries:
    (a) Local Query: The answer can be found in specific documents. Dense retriever ranks all documents and selects top-k, which contains the relevant information.
    (b) Global Query: The answer requires information from \textit{all} documents. However, dense retriever only returns top-k ranked documents, missing critical information scattered across the entire corpus.
    }
    \label{fig:local-global-rag-comparison}
\end{figure}

Retrieval-augmented generation (RAG) has been proposed as a solution to mitigate hallucination and knowledge limitations in large language models (LLMs) \cite{rag, r3-rag, aiocean, largefundataion, omni}. Its core mechanism lies in retrieving relevant chunks from a corpus to inject factual knowledge into the generation process, thereby improving the reliability of LLM outputs\cite{gao2023verifiable}.  

Current RAG systems operate at the \textbf{chunk level}, where the goal is to retrieve specific text segments from documents to answer queries. This approach are designed for answering questions that require \textbf{localized information}—for instance, determining when Einstein was born or identifying the capital of France. Current evaluation benchmarks exclusively focus on such local RAG tasks. Single-hop datasets like Natural Questions \cite{nq} and MS MARCO \cite{ms} test the extraction of facts from individual documents, while multi-hop datasets such as HotpotQA \cite{hotpotqa} and 2WikiMultihopQA \cite{2wiki} evaluate reasoning across a small number of connected documents.

However, many real-world applications require \textbf{corpus-level} operations that aggregate information across entire document collections, as show in figure \ref{fig:local-query-rag}. Systems for multi-document question answering \cite{multi-document} demonstrate the need for analytical capabilities that extract and combine facts across large document corpora. Recent OpenAI systems \cite{openai2024openaio1card} have been developed to conduct multi-step research across hundreds of sources, highlighting growing practical demand for such capabilities. As show in figre \ref{fig:global-query-rag}, these \textbf{global RAG} tasks—such as ``Which domain has the
highest average years of experience?''—cannot be answered by retrieving individual documents. Instead, they require traversing thousands of documents, extracting comparable attributes, and performing corpus-wide analysis. GraphRAG \cite{graphrag} shows that baseline RAG performs poorly on such global queries, yet no systematic benchmark exists to evaluate these corpus-level reasoning capabilities.

To address this gap, we construct \textbf{GlobalQA}, the first benchmark explicitly specifically to evaluate \textbf{global RAG} abilities. Since global tasks often require integrating and analyzing a large number of documents, GlobalQA defines four task types:  
- Counting: computing the number of entities that satisfy given conditions;  
- Extremum queries: identifying entities with extreme attributes;  
- Sorting: ranking entities on a global scale;  
- Top-$k$ extraction: retrieving the top-$k$ entities after ranking.  

Through systematic evaluation on GlobalQA, we find that existing RAG methods \cite{rag, ircot, hypergraphrag} achieve at most 1.51 F1 score on global tasks. We manually analyzed the error cases and identified three fundamental limitations that contribute to the poor performance:

\noindent\textbf{Issue 1: Fixed-granularity chunking disrupts document integrity.}  
Mainstream RAG systems split documents into fixed-length chunks (e.g., 512 tokens) \cite{rag}, which mechanically breaks the inherent structure of documents \cite{chunkfree}. For example, after chunking a research paper, metadata (e.g., publication year, conference name) and citation counts may be separated into different segments, preventing the system from correctly linking attributes to values, often causing double-counting or omissions.  

\noindent\textbf{Issue 2: Dense retrieval returns semantically relevant but factually irrelevant noise.}  
Dense retrievers frequently return many documents that are semantically related but do not contain the required answer. Prior work \cite{noiserag, wang2023retrieval} shows that such high-scoring yet irrelevant documents can harm LLM performance. For the above query, the retriever may return documents discussing ``citation analysis methods'' or ``AI conference overviews,'' but not the actual citation data. These noisy documents consume the limited context window and distract the model from attending to the truly relevant document.  

\noindent\textbf{Issue 3: Inherent limitations of LLMs in numerical computation.}  
Recent studies \cite{numerologic}demonstrate that LLMs generally perform poorly on numerical reasoning tasks. Even when provided with all necessary information, models often make mistakes in numerical comparisons, omit key items, or produce inconsistent results.  

To tackle the above challenges, we propose three technical innovations:  
\textbf{(1) Document-level retrieval:} using entire documents as retrieval units to preserve structural integrity and semantic coherence, ensuring that all relevant attributes can be processed jointly.  
\textbf{(2)Filter:} designing a retrieval–reading–filtering pipeline, where an LLM-driven filter eliminates noisy documents before reasoning, ensuring that inference is based on high-quality documents.  
\textbf{(3) Data processing tools:} introducing auxiliary tools to assist the model in handling numerical and statistical problems, enabling hybrid reasoning that combines language understanding with symbolic computation.  

The main contributions of this paper can be summarized as follows:  
\begin{enumerate}  
    \item First, we introduce GlobalQA, establishing the first systematic benchmark for evaluating global RAG capabilities.
    \item Second, through comprehensive experiments, we demonstrate that existing RAG approaches achieve only 1.51 F1 score on global tasks, revealing fundamental architectural limitations that cannot be addressed through incremental improvements.
    \item Third, we present GlobalRAG, which achieves 6.63 F1 score on GlobalQA with Qwen2.5-14B—a 5 point improvement over the strongest baseline.
\end{enumerate}  

The dataset is publicly available at \href{https://huggingface.co/datasets/QiiLuoo/GlobalQA/tree/main}{\texttt{GlobalQA}}.

\footnote{\url{https://huggingface.co/datasets/QiiLuoo/GlobalQA}}

\section{Related Work}

\subsection{Retrieval-Augmented Generation Methods}

The development of retrieval-augmented generation (RAG) can be broadly categorized into two paradigms: unstructured retrieval and structured retrieval.  

\paragraph{Unstructured Retrieval Methods:}

Unstructured retrieval directly indexes and searches over raw documents, and is the mainstream paradigm in current RAG systems.  
Dense document Retrieval (DPR)~\cite{dpr} leverages a dual-encoder architecture to learn dense representations of queries and documents, achieving success in open-domain QA.  
Contriever~\cite{contriever} further introduces an unsupervised contrastive learning framework to enhance the generalization ability of retrievers.  
RETRO~\cite{retro} integrates retrieval deeply into the Transformer architecture, achieving tighter coupling between retrieval and generation. Recent studies focus on optimizing retrieval strategies and reasoning processes.  
Self-RAG~\cite{self-rag} incorporates a self-reflection mechanism, using special tokens to control retrieval timing and verify retrieved content.  
FLARE~\cite{flare} proposes an active retrieval strategy that dynamically triggers retrieval based on generation confidence.  
IRCoT~\cite{ircot} interleaves chain-of-thought reasoning with retrieval, demonstrating advantages in multi-hop reasoning tasks.
Recent methods have begun exploring reinforcement learning to optimize multi-step retrieval processes \cite{research, search-r1}. 
However, these methods are primarily designed for optimizing local information retrieval and face fundamental limitations when large-scale information aggregation is required.  

\paragraph{Structured Retrieval Methods:}

Structured retrieval organizes information by constructing knowledge graphs or hierarchical indexes, aiming to capture semantic relationships between documents.  
KG-RAG~\cite{kg-rag} transforms unstructured text into knowledge triples and performs multi-hop reasoning through graph traversal.  
GraphRAG~\cite{graphrag} builds multi-level graph structures via community detection, marking the first attempt to address global query tasks.  
RAPTOR~\cite{raptor} constructs tree-structured indexes using recursive summarization, supporting multi-granularity retrieval. More recent graph-based approaches further enhance knowledge representation.  
HippoRAG~\cite{hipporag} simulates hippocampal memory mechanisms and optimizes retrieval paths through personalized PageRank.  
HyperGraphRAG~\cite{hypergraphrag} extends graph structures into hypergraphs, where hyperedges can connect multiple nodes to better model complex multi-relational structures.  
However, information loss during graph construction and the locality of graph traversal still limit their performance on global RAG tasks.  

\subsection{Evaluation Benchmarks for RAG}

\paragraph{Single-Hop Retrieval Datasets:}

Single-hop retrieval datasets\cite{nq,ms,triviaqa, popqa} mainly evaluate the system’s ability to locate a single relevant piece of information.  
Natural Questions (NQ)~\cite{nq} contains real queries from Google Search, where answers typically lie in a single Wikipedia document.  
MS MARCO~\cite{ms} provides 1M query-document pairs and has become the standard resource for training dense retrievers.  
TriviaQA~\cite{triviaqa} consists of 95K QA pairs, emphasizing factual knowledge retrieval.  
A common limitation of these datasets is that answers are restricted to a single document span, making them incapable of evaluating information aggregation.  
\paragraph{Multi-Hop Reasoning Datasets:}
Multi-hop datasets \cite{hotpotqa, 2wiki, musique}require systems to integrate information across multiple documents for reasoning.  
HotpotQA~\cite{hotpotqa} includes 113K questions requiring two-hop reasoning, with annotated supporting facts.  
2WikiMultihopQA~\cite{2wiki} ensures unique reasoning paths through template-based generation.  
MuSiQue~\cite{musique} extends to 4-hop reasoning and provides decomposed annotations from single-hop to 4-hop reasoning. However, existing multi-hop datasets still focus on \emph{multi-hop chains}, i.e., connecting a small set of documents through a limited reasoning sequence.  
In contrast, global RAG tasks require \emph{parallel aggregation} of information from a large number of documents.  
This qualitative difference prevents existing benchmarks from accurately reflecting a system’s ability to perform corpus-level reasoning.  
Table~\ref{tab:dataset-comparison} provides a detailed comparison between GlobalQA and existing datasets.  

\begin{figure*}
    \centering
    \includegraphics[width=1.0\linewidth]{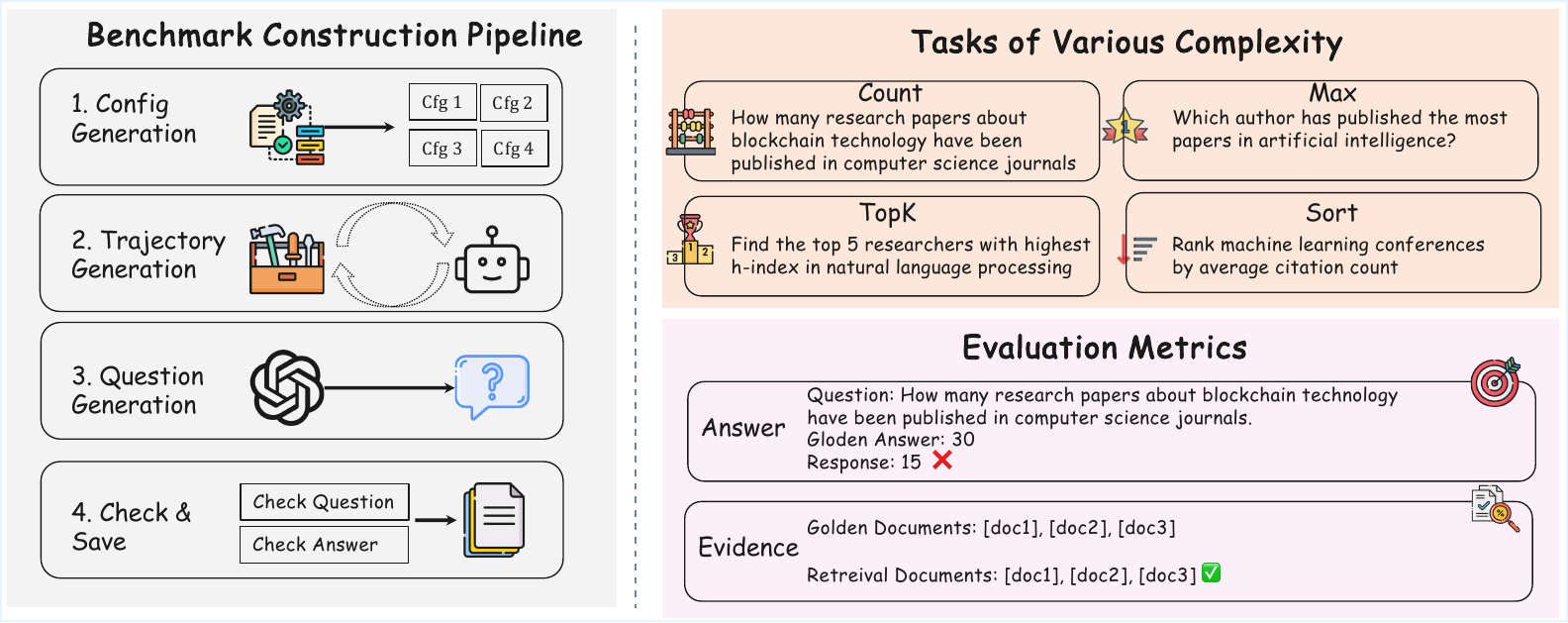}
    \caption{GlobalQA benchmark overview: construction pipeline (left), task examples with various complexity (top-right), and evaluation metrics (bottom-right).}    
    \label{fig:data-construction}
\end{figure*}


\section{The GlobalQA Dataset}

Existing retrieval-augmented QA datasets primarily focus on local, single-document retrieval tasks, making it difficult to assess a system’s ability to perform corpus-level information aggregation. To fill this gap, we introduce \textbf{GlobalQA}—a benchmark specifically designed to evaluate the corpus-level aggregation capability of retrieval-augmented systems. In this section, we present the dataset overview and construction pipeline, followed by a comparison with existing datasets and a discussion of evaluation metrics.  

\subsection{Overview}


\begin{figure*}[t]
    \centering
    \begin{subfigure}{0.32\linewidth}
        \centering
        \includegraphics[width=0.75\linewidth]{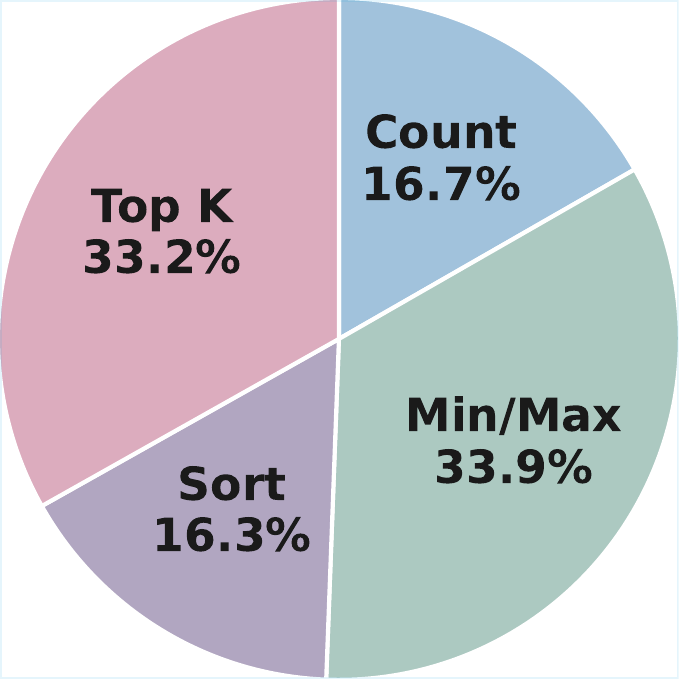}
    \end{subfigure}\hfill
    \begin{subfigure}{0.32\linewidth}
        \centering
        \includegraphics[width=\linewidth]{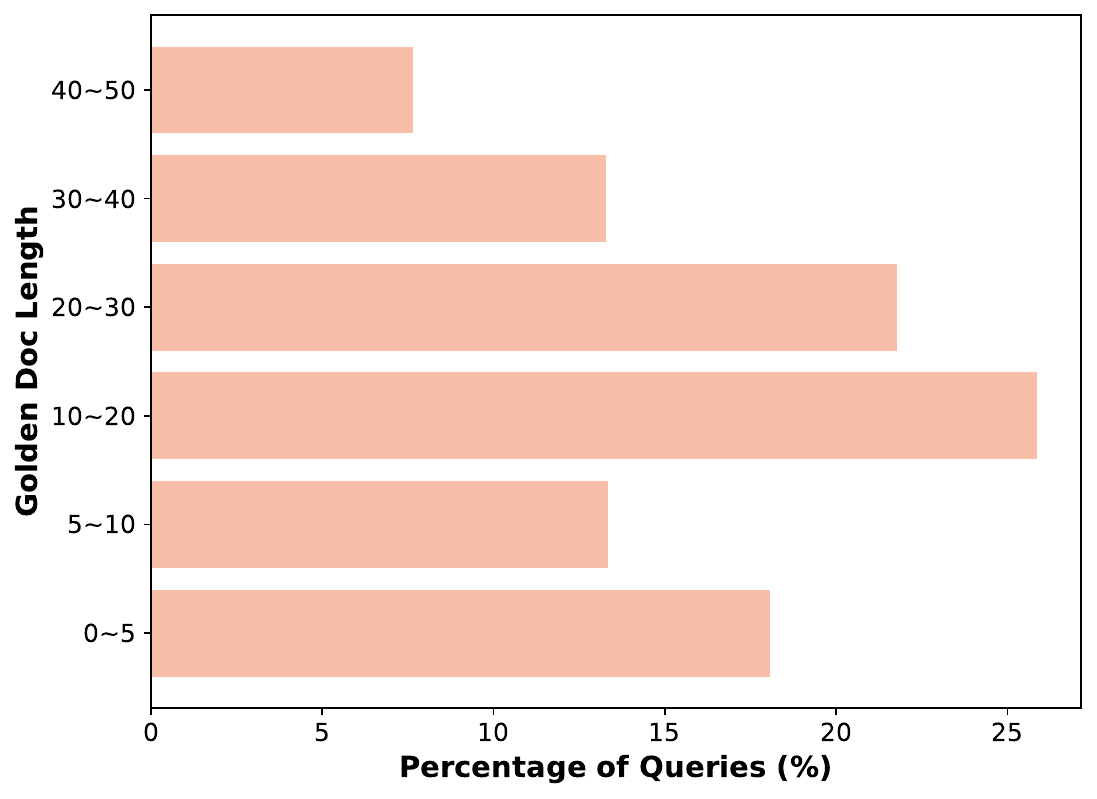}
    \end{subfigure}\hfill
    \begin{subfigure}{0.32\linewidth}
        \centering
        \includegraphics[width=\linewidth]{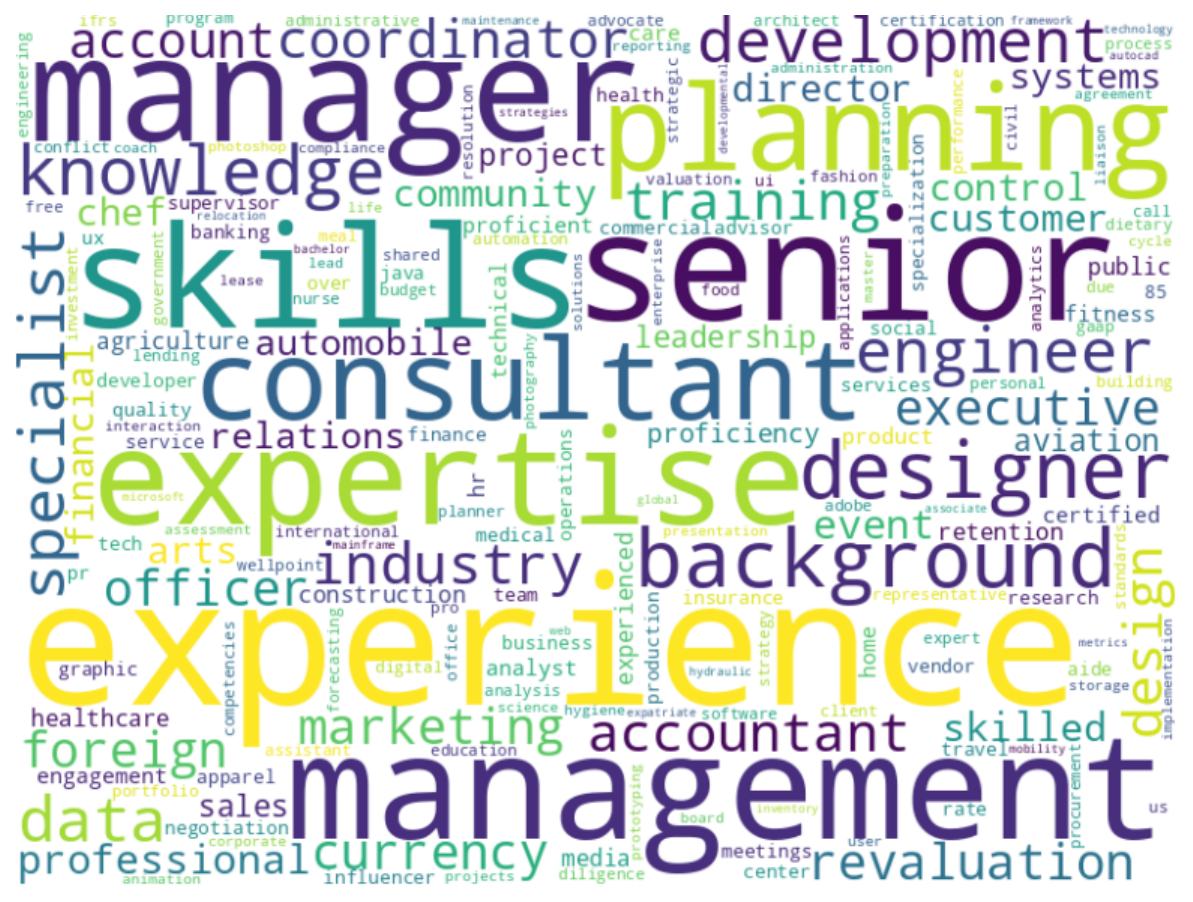}
    \end{subfigure}

    \caption{Statistical analysis of the GlobalQA dataset. 
    \textbf{Left}: distribution of task types. 
    \textbf{Middle}: distribution of the number of documents per query. 
    \textbf{Right}: keyword distribution.}
    \label{fig:threepdfs}
\end{figure*}

Table~\ref{fig:threepdfs} summarizes the dataset statistics. GlobalQA contains more than 13,000 question-answer pairs, constructed on a corpus of more than 2,000 real-world resumes across 23 professional domains. The dataset exhibits the following characteristics:  

\textbf{Task diversity}: GlobalQA covers four core task types—\textbf{Counting} (16.7\%), which computes the number of entities satisfying given conditions; \textbf{Extremum queries} (Max/Min, 33.9\%), which identify entities with extreme attributes; \textbf{Sorting} (16.3\%), which ranks entities globally; and \textbf{Top-$k$ extraction} (33.9\%), which retrieves the top-$k$ entities after sorting. These tasks comprehensively evaluate the model’s ability in statistics, comparison, ranking, and filtering.  

\textbf{Corpus-Level document Requirements}: Unlike existing benchmarks where answers typically reside in 1--4 documents, GlobalQA queries necessitate traversing significantly larger document sets. As shown in Figure~\ref{fig:threepdfs} (middle), 42.6\% of queries require documents from more than 20 documents, with the maximum reaching 50 documents per query. Specifically: 18.1\% require 2--5 documents, 13.4\% require 5--10, 25.9\% require 10--20, and the remaining 42.6\% demand over 20 documents. This distribution ensures that models cannot rely on heuristics designed for few-document scenarios and must develop genuine corpus-wide reasoning strategies.

\textbf{Keyword distribution}: Figure~\ref{fig:threepdfs} (right) visualizes the query keyword distribution. Query keywords are concentrated on high-frequency concepts such as ``experience'', ``management'', ``skills'', and ``project'' which are aligned with the structure of the resume corpus. The long-tail distribution of keywords covers both common scenarios and rare queries, allowing the dataset to evaluate model robustness in less frequent cases.

\subsection{Construction Pipeline}
Mainstream QA datasets\cite{hotpotqa, 2wiki} are typically constructed by starting from natural language questions and then manually annotating answers. This approach works for local retrieval tasks, where answers often reside in a small number of documents, making human verification feasible. However, global RAG tasks require analyzing and aggregating hundreds of documents for statistics, ranking, or set operations—making manual verification impractical and inconsistent.  

To address this challenge, GlobalQA adopts a \textbf{reverse construction strategy}: As shown in Figure~\ref{fig:data-construction}, we first programmatically design query trajectories, then allow an agent to execute these trajectories step by step to obtain deterministic answers, and finally generate natural language questions based on the completed trajectories. Since each trajectory execution is deterministic—guaranteed by the correctness of sub-trajectory inputs and outputs—the overall reasoning chain yields reliable answers, reducing the annotation requirements for large-scale corpus analysis.

The construction pipeline consists of three key stages:  

\textbf{Stage 1: Automatic configuration generation}  
The system automatically generates query configurations, including:  
- randomly sampling 2–5 trajectory steps to ensure diverse complexity,  
- selecting query domains from a predefined knowledge base,  
- designing retrieval steps combining keyword- and semantics-based queries,  
- generating set operations (intersection, union) for document aggregation,  
- assigning task types (counting, sorting, top-$k$ extraction),  
- converting configurations into natural language queries via templates.  

\textbf{Stage 2: Trajectory execution and data generation}  
\emph{Document retrieval and verification}: Based on the generated configuration, the system retrieves relevant documents from the corpus. To ensure accuracy and completeness, we employ the DeepSeek model to traverse the entire corpus, validating and supplementing documents potentially missed by the retriever. This ensures each trajectory is grounded in a complete and accurate document set.  
\emph{Trajectory data generation}: The system applies predefined set operations, computes relevance scores, executes counting or sorting tasks, assembles structured trajectory data, and filters out low-quality trajectories with abnormal document counts.  

\textbf{Stage 3: Natural language generation}  
Structured trajectories are converted into natural language questions by parsing key information and logical relations. Standard answers are extracted from execution results, ensuring full consistency among questions, trajectories, and answers.  

\textbf{Stage 4: Check and Save}
To control query difficulty and ensure dataset quality, we filter out queries requiring more than 50 documents. Additionally, we remove trajectories with abnormal document distributions or execution inconsistencies, ensuring all retained question-answer pairs have reliable ground truth and consistent reasoning paths.

\begin{table*}[t]
\centering
\small
\begin{tabular}{lcccc}
\toprule
\textbf{Dataset} & \textbf{Info Requirement} & \textbf{Docs} & \textbf{Hops} & \textbf{Example} \\
\midrule
\multicolumn{5}{l}{\textit{Single-hop datasets}} \\
NQ & chunk-level & 1 & Single & ``When was Einstein born?'' \\
MS MARCO & chunk-level & 1 & Single & ``What is photosynthesis?'' \\
TriviaQA & chunk-level & 1 & Single & ``Capital of France?'' \\
\midrule
\multicolumn{5}{l}{\textit{Multi-hop datasets}} \\
HotpotQA & chunk-level & 2--10 & Multi & ``When was director X's wife born?'' \\
2WikiMultihop & chunk-level & 2--4 & Multi & ``Where did A's CEO graduate?'' \\
MuSiQue & chunk-level & 2--4 & Multi & ``How old was Y when X occurred?'' \\
\midrule
\multicolumn{5}{l}{\textit{Global aggregation dataset}} \\
\textbf{GlobalQA} & \textbf{Corpus-level} & \textbf{2--50} & \textbf{Both} & \textbf{``Which domain has longest avg experience?''} \\
\bottomrule

\end{tabular}
\caption{Core differences between GlobalQA and existing RAG evaluation datasets.}
\label{tab:dataset-comparison}
\end{table*}

\subsection{Comparison with Existing Benchmarks}
Table~\ref{tab:dataset-comparison} highlights the differences between GlobalQA and existing RAG evaluation datasets. Existing datasets fall into two categories: \emph{single-hop retrieval datasets} (e.g., NQ, MS MARCO, TriviaQA), which mainly test single-document retrieval and simple factoid queries; and \emph{multi-hop datasets} (e.g., HotpotQA, 2WikiMultihop, MuSiQue), which involve multiple documents but remain limited to path-chaining or compositional reasoning over a small set of documents.  

In contrast, GlobalQA focuses on queries requiring comprehensive corpus-level analysis and global pattern recognition. Each query averages 2–50 documents, involving both single-hop and multi-hop reasoning. Typical tasks, such as \emph{``Which domain has the longest average experience?''}, demand statistical analysis over the entire corpus—capabilities that existing datasets cannot evaluate.

\subsection{Evaluation Metrics}

We adopt two metrics to evaluate performance:  

\textbf{F1}: measures the quality of final answers, using the standard token-level F1 score:  
\begin{equation}
\text{F1} = 2 \cdot \frac{\text{Precision} \cdot \text{Recall}}{\text{Precision} + \text{Recall}}
\end{equation}

\textbf{Document F1@k(D-F1@k)}: evaluates the coverage of retrieved documents, defined as the F1 score between the retrieved document set and the gold document set:  
\begin{equation}
\text{Precision@k} = \frac{|\mathcal{D}\text{ret}^k \cap \mathcal{D}\text{gold}|}{|\mathcal{D}_\text{ret}^k|}
\end{equation}\begin{equation}
\text{Recall@k} = \frac{|\mathcal{D}\text{ret}^k \cap \mathcal{D}\text{gold}|}{|\mathcal{D}_\text{gold}|}
\end{equation}\begin{equation}
\text{F1@k} = \frac{2 \cdot \text{Precision@k} \cdot \text{Recall@k}}{\text{Precision@k} + \text{Recall@k}}
\end{equation}




\section{Method}

Existing RAG methods face three key challenges in global RAG tasks: (i) document chunking disrupts structural integrity, (ii) noisy retrievals occupy the limited context window, and (iii) LLMs exhibit limited ability in numerical computation. To address these issues, we propose \textbf{GlobalRAG}, a multi-tool collaborative training-free framework

\subsection{Framework Overview}

GlobalRAG operates through a three-stage pipeline: \emph{retrieval → filtering → aggregation}. Given a global query, the system first retrieves complete documents rather than fragmented chunks, then applies LLM-driven filters to remove irrelevant noise, and finally invokes task-specific aggregation tools to derive accurate answers. 

\subsection{Document-Level Retrieval}
To preserve document integrity, GlobalRAG treats entire documents as atomic retrieval units rather than arbitrary text chunks.

\paragraph{Index construction}: For each document $D_i$ in the corpus, we build a chunk-level index:
\begin{equation}
\mathcal{I} = \{(D_1, \text{emb}(D_1)), ..., (D_n, \text{emb}(D_n))\}
\end{equation}

\paragraph{Retrieval strategy}: Given a query $q$, the retriever returns the top-$k$ most similar documents:
\begin{equation}
\mathcal{D}_{\text{ret}} = \text{TopK}(\text{sim}(\text{emb}(q), \text{emb}(D_i)), k)
\end{equation}

This preserves document structure and metadata.

\subsection{Document-Level Filter}

Dense retrievers return semantically relevant but task-irrelevant documents, introducing substantial noise. To mitigate this, we design a two-stage filtering mechanism. A lightweight LLM is used to precisely determine document relevance:
\begin{equation}
\begin{aligned}
r_i = \operatorname{LLM}_{\text{filter}}\Bigl(&\text{``Does document } D_i \\ \text{ contain information}
&\text{ to answer query } q \text{?''}\Bigr)
\end{aligned}
\end{equation}

This step discards irrelevant documents before reasoning, ensuring that inference is based on highly related documents.  

\subsection{Task-Level aggregation Tools} 
LLMs exhibit systematic deficiencies in large-scale numerical computation, statistical analysis, and precise ranking. When handling corpus-level queries that involve dozens or even hundreds of documents, models often suffer from counting errors, inconsistent numerical comparisons, or unstable sorting. To support numerical computation in global RAG, we design specialized computation modules that leverage symbolic precision to complement neural reasoning.  

\textbf{Tool design}: We introduce four specialized tools, each tailored to specific global RAG needs.  
- The \textbf{Counting tool} systematically traverses documents, applies deduplication strategies, and accurately enumerates entities for corpus-level counting tasks.  
- The \textbf{Extremum tool} extracts numerical and ordinal attributes, performing exact comparisons without approximation errors to resolve global optimization queries.  
- The \textbf{Sorting tool} extracts comparable metrics from heterogeneous document formats, accounts for scale differences across documents, and guarantees consistent results through deterministic algorithms.  
- The \textbf{Top-$k$ extraction tool} leverages efficient heap-based algorithms with pruning strategies to identify the top-$k$ entities, balancing global analysis with selective output.  

These four tools can be combined to form a computation framework. Together, they provide a reliable foundation for numerical reasoning in GlobalRAG, ensuring both accuracy and consistency in global RAG tasks.  


\section{Experimental Setup}

\subsection{Baseline Methods}
To comprehensively evaluate the performance of different retrieval-augmented paradigms on global RAG tasks, we select five representative baseline methods, covering three major technical directions: single-shot retrieval, iterative retrieval, and graph-structured retrieval.  
\paragraph{Single-Shot Retrieval Methods:} StandardRAG~\cite{rag}: The standard retrieval-augmented generation approach, which performs a single round of dense retrieval to obtain the top-$k$ relevant documents as context.  
\paragraph{Iterative Retrieval Methods:}
FLARE~\cite{flare}: Dynamically triggers retrieval by monitoring uncertainty during generation. When the model produces tokens with low confidence, generation is paused and additional retrieval is performed. The confidence threshold is set to 0.5.IRCoT~\cite{ircot}: Integrates chain-of-thought reasoning with interleaved retrieval, where each reasoning step triggers a retrieval operation. While suitable for multi-step reasoning tasks, its rigid structure may limit performance in scenarios requiring parallel information aggregation.  
\paragraph{Graph-Structured Retrieval Methods:}
HippoRAGv2~\cite{hipporag}: Inspired by hippocampal memory mechanisms, this method constructs a knowledge graph and applies personalized PageRank to optimize retrieval paths. A ``memory signal'' mechanism is introduced to enhance adaptation to historical query patterns. HyperGraphRAG~\cite{hypergraphrag}: Extends graph structures into hypergraph representations. It builds multi-granularity community structures through entity extraction, relation modeling, and hierarchical aggregation, combining local neighborhood expansion with global community matching for retrieval.  

\subsection{Implementation Details}
During dataset construction, we employ DeepSeek-v3 to validate the completeness of retrieved documents for each query trajectory. For evaluation, we use Qwen2.5-Instruct models as the backbone LLM. The retriever is BGE-large-en, and the filter module uses Qwen3-4B. We set the maximum retrieval iterations to 10 and retrieve top-20 documents per query. Temperature is set to 0. All baseline methods use identical retriever and LLM configurations for fair comparison.


\section{Main Experiments}

Table~\ref{tab:main} presents the performance comparison between our proposed \textbf{GlobalRAG} and multiple baseline methods across different Qwen2.5 model scales (3B, 7B, 14B) \cite{qwen2.5} and four categories of global RAG tasks. The results show that GlobalRAG consistently outperforms all baselines in terms of average F1 score. On the 14B model, GlobalRAG achieves an average F1 of 6.63 and an average D-F1@20 of 12.01, surpassing the strongest baseline, StandardRAG, by 5.12 and 3.92 points, respectively.  

\paragraph{Comparison with Iterative Retrieval Methods.}  
On the 14B model, GlobalRAG’s average F1 score is 6.54 points higher than IRCoT and 6.08 points higher than FLARE. This gap correlates with the predefined retrieval–reasoning workflows of iterative methods, which were originally designed for single-hop or multi-hop factoid QA. Such rigid pipelines constrain flexibility in global RAG scenarios, where the system must dynamically adjust its information acquisition strategy. In contrast, GlobalRAG empowers the model with tool invocation mechanisms, enabling it to autonomously select and combine external resources according to task requirements, thereby unleashing the full reasoning potential of LLMs. Notably, although IRCoT achieves relatively high D-F1@20 scores (e.g., 10.38 on the 3B model), indicating that it can retrieve more relevant documents, its answer F1 remains much lower than GlobalRAG (1.00 point lower on 3B and 6.54 points lower on 14B). This highlights that merely retrieving more relevant information is insufficient for global RAG tasks—effective integration and processing of the retrieved document are equally crucial. Constrained by its iterative mode, IRCoT struggles to perform global synthesis over dispersed information.  

\paragraph{Comparison with Graph-Structured Methods.}  
As a structured method that transforms documents into knowledge graphs, HyperGraphRAG performs poorly on global RAG tasks. Results show that its F1 score does not exceed 0.47 across all model scales, and its D-F1@20 score cannot even be computed (due to failure in returning valid document documents). By contrast, GlobalRAG achieves an average F1 that is 6.54 points higher than HyperGraphRAG on the 14B model. The performance gap correlates with information loss during graph construction. HyperGraphRAG decomposes each document into sets of nodes and edges, establishing relational networks between entities but destroying the structural integrity of documents as unified information units. In global RAG tasks, chunk-level metadata (e.g., document counts, attribute distributions) is crucial, but such information is lost during graph transformation. For instance, in counting tasks, systems must tally the number of documents satisfying certain conditions, yet HyperGraphRAG cannot trace original document boundaries, leading to failure. By contrast, GlobalRAG preserves document-level retrieval, ensuring structural integrity and allowing the model to directly access and manipulate complete document units, thereby achieving breakthrough performance in global RAG tasks.  

\paragraph{Cross-Scale Performance Trends.}  
As model size increases from 3B to 14B, GlobalRAG shows substantial improvement (average F1 rises from 2.52 to 6.63, a gain of 4.11 points), while other methods exhibit limited or unstable improvements. This demonstrates that GlobalRAG’s multi-tool collaborative framework better leverages the enhanced reasoning capacity of larger LLMs, enabling effective scalability in performance.

\begin{table*}[]
\small
\centering
\resizebox{0.93\textwidth}{!}{%
\begin{tabular}{lrrrrrrrrrr}
\toprule
Task & \multicolumn{2}{c}{TopK} & \multicolumn{2}{c}{Count} & \multicolumn{2}{c}{Sort} & \multicolumn{2}{c}{MinMax} & \multicolumn{2}{c}{Avg} \\
\cmidrule(lr){2-3} \cmidrule(lr){4-5} \cmidrule(lr){6-7} \cmidrule(lr){8-9} \cmidrule(lr){10-11}
Metric & D-F1@20 & F1 & D-F1@20 & F1 & D-F1@20 & F1 & D-F1@20 & F1 & D-F1@20 & F1 \\
\midrule
\multicolumn{11}{c}{Qwen2.5-3B-Instruct} \\
StandardRAG & 7.98 & \textbf{2.14} & 8.88 & 0.93 & 7.65 & 2.58 & 7.99 & 0.80 & 8.09 & 1.56 \\
FLARE & 0.01 & 0.18 & 0.00 & \textbf{2.33} & 0.03 & 0.10 & 0.00 & 0.53 & 0.01 & 0.68 \\
IRCOT & \textbf{10.01} & 1.67 & 10.30 & 1.09 & 11.03 & 1.56 & \textbf{10.44} & \textbf{1.60} &\textbf{10.38} & 1.52 \\
HyperGraphRAG & - & 0.06 & - & 0.39 & - & 0.04 & - & 0.00 & - & 0.09 \\
\rowcolor{gray!10} GlobalRAG (ours) & 2.14 & 1.32 &  \textbf{13.95} & 0.02 &  \textbf{12.95} &  \textbf{9.18} & 3.35 & 1.51 &  6.62 & \textbf{2.52} \\
\midrule
\multicolumn{11}{c}{Qwen2.5-7B-Instruct} \\
StandardRAG & \textbf{7.98} & 1.64 & 8.88 & 0.00 & 7.65 & 2.73 & 7.99 & 1.33 & 8.09 & 1.43 \\
FLARE & 0.33 & 0.00 & 0.00 & 2.33 & 0.30 & 0.11 & 0.53 & 0.53 & 0.33 & 0.62 \\
IRCOT & 7.76 & 1.30 & 8.68 & 0.04 & 7.68 & 0.96 & \textbf{8.25} & \textbf{1.34} & \textbf{8.07} & 1.02 \\
HyperGraphRAG & - & 0.40 & - & 0.46 & - & 0.23 & - & 0.00 & - & 0.25 \\
\rowcolor{gray!10} GlobalRAG (ours) & 2.65 & \textbf{1.75} & \textbf{9.31} & \textbf{2.79} & \textbf{10.45} & \textbf{9.13} & 1.50 & 0.80 & 4.89 & \textbf{2.93} \\
\midrule
\multicolumn{11}{c}{Qwen2.5-14B-Instruct} \\
StandardRAG & 7.98 & 1.78 & 8.88 & 0.93 & 7.65 & 1.96 & 7.99 & 1.33 & 8.09 & 1.51 \\
FLARE & 0.04 & 0.00 & 0.00 & \textbf{2.33} & 0.05 & 0.17 & 0.10 & 0.27 & 0.05 & 0.55 \\
IRCOT & 8.86 & 0.13 & 8.52 & 0.01 & 9.56 & 0.25 & 8.38 & 0.00 & 8.77 & 0.09 \\
HyperGraphRAG & - & 0.01 & - & 0.47 & - & 0.04 & - & 0.00 & - & 0.09 \\
\rowcolor{gray!10} GlobalRAG (ours) & \textbf{11.28} & \textbf{7.92} & \textbf{12.92} & 0.01 & \textbf{15.01} & \textbf{14.65} & \textbf{10.55} & \textbf{4.80} & \textbf{12.01} & \textbf{6.63} \\
\bottomrule
\end{tabular}
}
\caption{Performance comparison across four aggregation tasks (TopK, Count, Sort, MinMax) and their average (Avg). We report Answer F1 (F1) and document F1 (D-F1@20) for different RAG methods using three Qwen2.5-Instruct model sizes.}
\label{tab:main}
\end{table*}

\section{Analysis Experiments}

\subsection{Ablation Study}
To systematically evaluate the contribution of each component, we conducted three ablation experiments: (1) replacing chunk-level retrieval with 512-token chunk-based retrieval, (2) removing the LLM-driven filter module while retaining chunk-level retrieval and aggregation tools, and (3) removing both filter and aggregation tool modules to create a chunk-level StandardRAG baseline.Table~\ref{tab:ablation} presents the detailed results.

\paragraph{Impact of retrieval granularity:} We first replaced the chunk-level retriever with a conventional chunk-based retriever. The results show that this replacement caused the F1 score to drop sharply from 7.27 to 0.4, a decrease of 94.5\%. This performance degradation indicates the contribution of chunk-level retrieval in global RAG tasks. Notably, since chunk-based retrieval cannot provide chunk-level boundary information, we did not compute statistics in this setting.

\paragraph{Progressive contribution of tool modules:} On top of chunk-level retrieval, we further analyzed the impact of each tool module. Removing the filter module led to drops of 0.53 and 1.72 in F1 and D-F1@20, corresponding to relative declines of 7.3\% and 14.1\%, respectively. This indicates that the filter removes redundant information and improves reasoning precision.

\paragraph{Necessity of a complete toolchain:} When both the filter and aggregation tool modules were removed, the model degenerated into a standard RAG framework. The performance dropped substantially to F1=1.51 and D-F1@20=8.09, corresponding to relative declines of 79.2\% and 33.6\% compared to the full framework. This result demonstrates the contribution of our tool-based design, especially the critical synergy between the filter and aggregation tools in supporting complex global RAG.

\begin{table}[ht]
\centering
\renewcommand{\arraystretch}{1.2}
\setlength{\tabcolsep}{8pt}
\begin{tabular}{lrr}
\hline
\textbf{Setting} & \textbf{F1} & \textbf{D-F1@20} \\
\hline
GlobalRAG & 7.27 & 12.18 \\
~~~w/ Chunk-based retrieval  & 0.40 & - \\
~~~w/o Filter & 6.74 & 10.46 \\
~~~~~~w/o aggregation module & 1.51 & 8.09 \\
\hline
\end{tabular}
\caption{Ablation study results on different components.}
\label{tab:ablation}
\end{table}

\subsection{Cross-Dataset Evaluation}
We evaluate representative methods on 2WikiMultihopQA, HotpotQA, MusiQue and GlobalQA using Qwen2.5-7B-Instruct.

Table~\ref{tab:cross-dataset} shows a striking performance gap. While methods achieve 17-41 F1 on 2WikiMultihopQA, HotpotQA benchmarks, all collapse below 2 F1 on GlobalQA—a 95-99\% relative decline. Notably, even HyperGraphRAG, designed for graph-structured reasoning, drops from 18.25 to 0.09 F1.

This reveals that multi-hop reasoning (chaining 2-4 documents) and global aggregation (parallel analysis of 20+ documents) are qualitatively different tasks. Existing retrieval paradigms—whether iterative (IRCoT, FLARE) or graph-based (HyperGraphRAG)—fail because they lack mechanisms for: (1) preserving document-level integrity during chunking, (2) filtering large-scale retrieval noise, and (3) performing corpus-wide statistical operations. This motivates our GlobalRAG framework.

\begin{table*}[ht]
\centering
\small
\begin{tabular}{lcccc}
\toprule
\textbf{Method} & \textbf{2Wiki} & \textbf{HotpotQA}& \textbf{MuSiQue}   & \textbf{GlobalQA}\\
\midrule
StandardRAG & 12.75 &  16.58 & 4.53 & 1.51 \\
HyperGraphRAG & 21.10 & 37.50 & 20.4 &0.09\\
FLARE & 28.30 & 24.80 & 2.80 &0.62 \\
IRCoT & 33.50 & 41.10 & 8.90  &1.02 \\
\bottomrule
\end{tabular}
\caption{Cross-dataset performance comparison using F1 score on Qwen2.5-7B-Instruct. Methods achieve reasonable performance on multi-hop QA benchmarks (2WikiMultihopQA, HotpotQA, MuSiQue) but experience catastrophic collapse on corpus-level reasoning tasks (GlobalQA), revealing a fundamental gap between local RAG and global RAG.}
\label{tab:cross-dataset}
\end{table*}

\subsection{Effect of Different Retrieval Steps}
Multi-step retrieval is a core mechanism of our method, essentially enabling iterative information gathering and integration to construct a more comprehensive global view. The number of retrieval steps directly affects the system's depth of understanding of complex information structures: too few steps may result in insufficient information collection and incomplete reasoning chains, while an appropriate increase allows our progressive reasoning framework to function more effectively. However, too many steps may run into the context window limitation of the model. Therefore, we systematically analyze the effect of varying retrieval steps on global RAG performance. Under the same experimental conditions, we gradually increased the maximum retrieval steps from 0 to 20 and evaluated GlobalRAG against the IRCOT baseline. All other hyperparameters were kept constant for a fair comparison.

\begin{figure}
    \centering
    \includegraphics[width=1\linewidth]{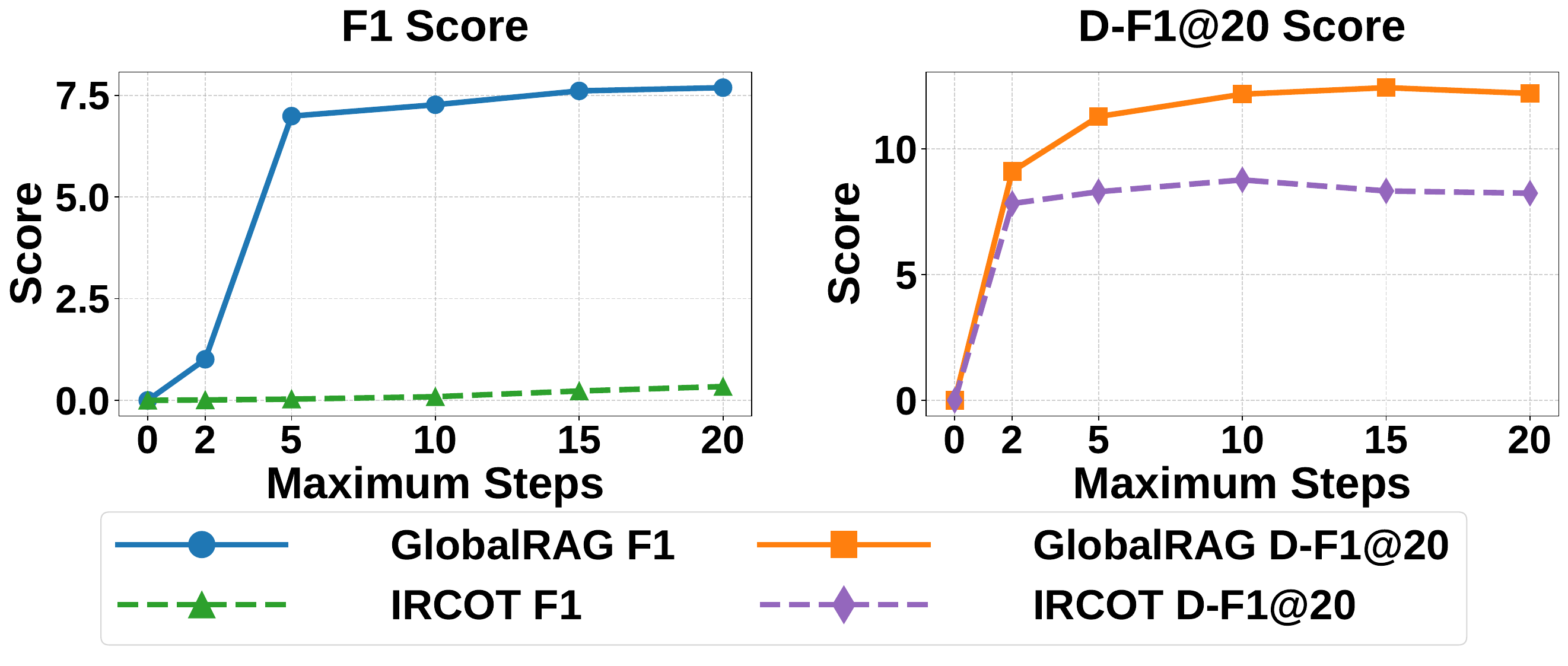}
    \caption{F1/D-F1@20 trends of GlobalRAG and IRCOT (baseline) under different retrieval steps.}
    \label{fig:different-steps}
\end{figure}

As shown in Figure~\ref{fig:different-steps}, both methods exhibited increasing F1 and D-F1@20 with more retrieval steps, but GlobalRAG consistently maintained a clear performance advantage. Specifically, in the 0–5 step range, GlobalRAG's F1 rose from nearly 0 to 7.0 and its D-F1@20 from 0 to 11.3, whereas IRCOT improved by less than 1 point. In the 5–20 step range, GlobalRAG continued to steadily improve, while IRCOT's growth flattened, and the performance gap widened. These results indicate that multi-step retrieval indeed brings gains, and our progressive reasoning framework is better able to leverage additional retrieval steps to build a global knowledge graph.

\subsection{Effect of Different Retrievers}
The success of global RAG depends heavily on whether the retriever can accurately fetch complete and relevant external information. We therefore examined the effect of retrievers of different parameter scales. We compared four retrievers: the BGE baseline, Qwen3-Embedding-0.6B, Qwen3-Embedding-4B, and Qwen3-Embedding-8B. Results in Table~\ref{tab:qwen-models} show that larger retrievers consistently provide higher-quality information and thus yield better global RAG accuracy. Using the BGE retriever as a baseline (F1=7.27, D-F1@20=12.18), Qwen3-0.6B brought improvements of 2.5\% and 0.7\%, while Qwen3-8B achieved gains of 18.7\% and 6.6\%. These results show that our framework maintains performance across different retriever sizes and its scalability: as retrievers improve, our method can further capitalize on the higher-quality documents to achieve larger performance gains. This indicates strong potential for future advances in global RAG with more powerful retrievers.

\begin{table}[ht]
\centering
\renewcommand{\arraystretch}{1.2}
\setlength{\tabcolsep}{8pt}
\begin{tabular}{lcc}
\hline
\textbf{Retriever} & \textbf{F1} & \textbf{D-F1@20} \\

\hline
bge-large-en & 7.27 & 12.18 \\
Qwen3-Embedding-0.6B & 7.45 & 12.27 \\
Qwen3-Embedding-4B   & 7.83 & 12.77 \\
Qwen3-Embedding-8B   & 8.63 & 12.99 \\
\hline
\end{tabular}
\caption{Performance of different retrievers on the dataset.}
\label{tab:qwen-models}
\end{table}

\subsection{Effect of Different Filters}
A key limitation of existing RAG methods in global RAG is their vulnerability to noisy documents. Our GlobalRAG framework addresses this by incorporating an intelligent filter that accurately identifies and removes irrelevant documents. The scale of the filter directly determines its precision and thus the overall reasoning quality. We compared three filters of different parameter sizes: Qwen3-4B, Qwen3-8B, and Qwen3-14B. Each filter was tested with the same chunk-level retriever and aggregation module to validate the effectiveness of our retrieval–reading–filter pipeline.

Table~\ref{tab:qwen-filter} shows that larger filters consistently improved GlobalRAG performance: Qwen3-4B achieved F1=7.27 and D-F1@20=12.18; Qwen3-8B yielded gains of 2.2\% and 0.9\%; and Qwen3-14B further increased scores to 7.79 and 12.46, representing relative improvements of 7.2\% and 2.3\%. Larger filters also reduced performance variance, improving stability. These findings confirm that LLM-driven filters effectively address noise interference, ensuring that only high-quality documents are passed to the aggregation execution module, thus enabling hybrid reasoning that combines language understanding with symbolic computation.

\begin{table}[ht]
\centering
\renewcommand{\arraystretch}{1.2}
\setlength{\tabcolsep}{8pt}
\begin{tabular}{lcc}
\hline
\textbf{Filter} & \textbf{F1} & \textbf{D-F1@20} \\
\hline
Qwen3-4B   & 7.27 & 12.18 \\
Qwen3-8B   & 7.43 & 12.29 \\
Qwen3-14B  & 7.79 & 12.46 \\
\hline
\end{tabular}
\caption{Performance of different filter sizes.}
\label{tab:qwen-filter}
\end{table}

\subsection{Effect of Different Retrieval Numbers}
\begin{figure}
    \centering
    \includegraphics[width=1\linewidth]{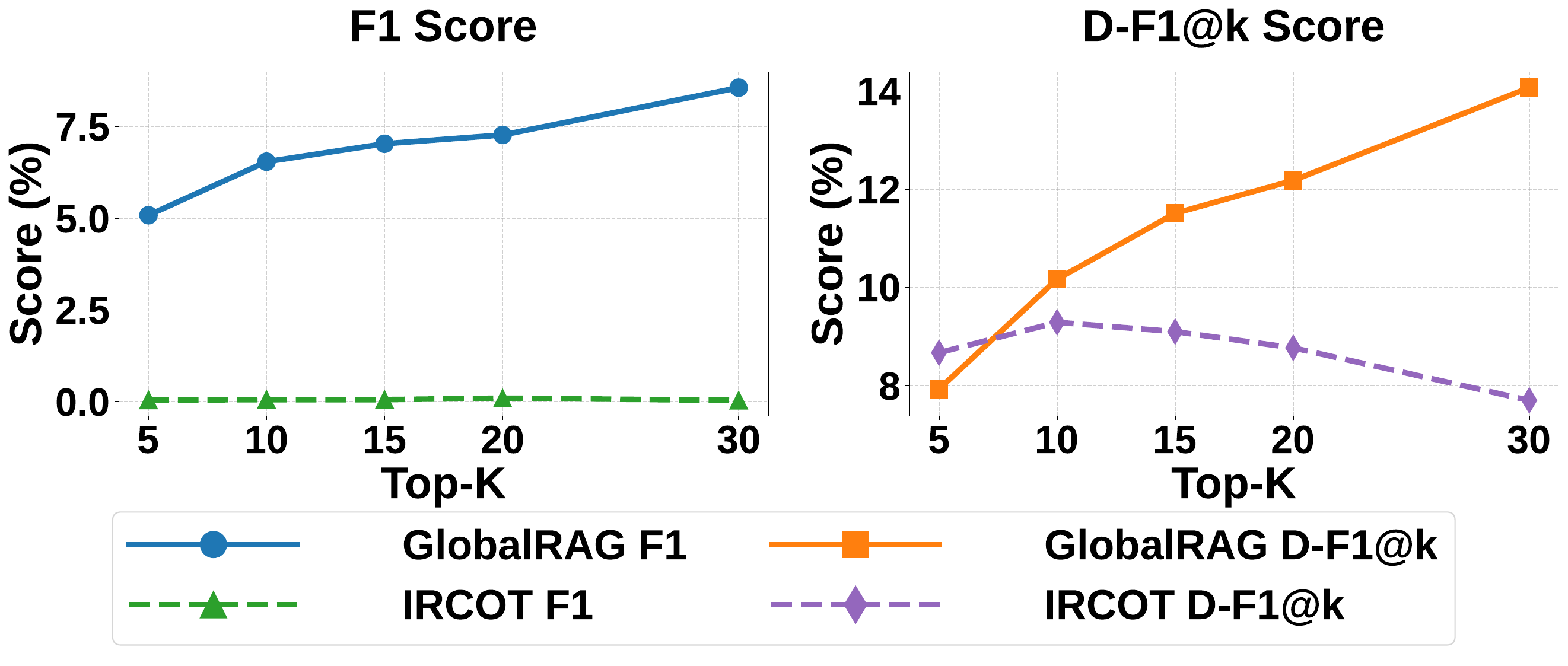}
    \caption{F1/D-F1@k trends of GlobalRAG and IRCOT (baseline) under different retrieval numbers (Top-K).}
    \label{fig:different-topk}
\end{figure}

In global RAG tasks, increasing the number of retrieved documents can introduce noise, while decreasing it risks missing key documents. We therefore evaluated the effect of varying retrieval numbers from 5 to 30, comparing GlobalRAG with IRCOT under the same retriever setup. As shown in Figure~\ref{fig:different-topk}, GlobalRAG’s F1 and D-F1@k steadily increased with more retrieved documents, reaching peak performance at 30 documents. This demonstrates that GlobalRAG’s design effectively selects relevant documents from larger candidate sets without being distracted by noise. Its filtering module enables it to leverage expanded recall to boost coverage while discarding irrelevant content.

In contrast, IRCOT peaked at 10 retrieved documents but then declined, with performance at 30 documents falling below that at 5. This highlights its lack of noise filtering: as more documents are retrieved, semantically related but factually irrelevant noise is passed into the model, crowding out relevant document and misleading reasoning, thus degrading performance.

\section{Conclusion}
This paper investigates the challenges of global query processing in retrieval-augmented generation systems. We make the following contributions:

First, we introduce GlobalQA, a novel benchmark specifically designed to evaluate RAG systems on global aggregation tasks. Unlike existing benchmarks that focus on local information retrieval, GlobalQA requires systems to perform corpus-wide traversal and multi-document reasoning, better reflecting real-world knowledge-intensive applications. Our benchmark comprises four aggregation task types that systematically assess reasoning capabilities across different computational complexities.

Second, through extensive empirical evaluation, we identify and quantify three fundamental limitations of current RAG architectures when handling global queries: (1) information fragmentation across distributed documents, (2) retrieval noise amplification in large-scale searches, and (3) computational bottlenecks in exhaustive corpus processing. Our analysis reveals that state-of-the-art RAG systems achieve only 1.5\% average F1 score on global tasks, highlighting a critical gap in current methodologies.

Third, we propose GlobalRAG, a hybrid architecture that demonstrates the necessity of multi-paradigm integration for complex reasoning tasks. By combining neural retrieval with programmatic execution, our approach achieves a 5\% performance improvement over existing methods, validating that neither pure neural nor pure symbolic approaches alone are sufficient for global query processing.

Our work identifies global query processing as a challenge in RAG research, providing both evaluation tools and architectural insights for advancing the field.

\bibliography{arXiv/custom}

\begin{thebibliography}{33}
\providecommand{\natexlab}[1]{#1}

\bibitem[{Asai et~al.(2023)Asai, Wu, Wang, Sil, and Hajishirzi}]{self-rag}
Akari Asai, Zeqiu Wu, Yizhong Wang, Avirup Sil, and Hannaneh Hajishirzi. 2023.
\newblock \href {https://arxiv.org/abs/2310.11511} {Self-rag: Learning to retrieve, generate, and critique through self-reflection}.
\newblock \emph{Preprint}, arXiv:2310.11511.

\bibitem[{Bajaj et~al.(2018)Bajaj, Campos, Craswell, Deng, Gao, Liu, Majumder, McNamara, Mitra, Nguyen, Rosenberg, Song, Stoica, Tiwary, and Wang}]{ms}
Payal Bajaj, Daniel Campos, Nick Craswell, Li~Deng, Jianfeng Gao, Xiaodong Liu, Rangan Majumder, Andrew McNamara, Bhaskar Mitra, Tri Nguyen, Mir Rosenberg, Xia Song, Alina Stoica, Saurabh Tiwary, and Tong Wang. 2018.
\newblock \href {https://arxiv.org/abs/1611.09268} {Ms marco: A human generated machine reading comprehension dataset}.
\newblock \emph{Preprint}, arXiv:1611.09268.

\bibitem[{Borgeaud et~al.(2022)Borgeaud, Mensch, Hoffmann, Cai, Rutherford, Millican, van~den Driessche, Lespiau, Damoc, Clark, de~Las~Casas, Guy, Menick, Ring, Hennigan, Huang, Maggiore, Jones, Cassirer, Brock, Paganini, Irving, Vinyals, Osindero, Simonyan, Rae, Elsen, and Sifre}]{retro}
Sebastian Borgeaud, Arthur Mensch, Jordan Hoffmann, Trevor Cai, Eliza Rutherford, Katie Millican, George van~den Driessche, Jean-Baptiste Lespiau, Bogdan Damoc, Aidan Clark, Diego de~Las~Casas, Aurelia Guy, Jacob Menick, Roman Ring, Tom Hennigan, Saffron Huang, Loren Maggiore, Chris Jones, Albin Cassirer, and 9 others. 2022.
\newblock \href {https://arxiv.org/abs/2112.04426} {Improving language models by retrieving from trillions of tokens}.
\newblock \emph{Preprint}, arXiv:2112.04426.

\bibitem[{Chen et~al.(2025)Chen, Sun, Li, Sun, Zhou, Zhu, Wang, Pan, Zhang, Chen, Yang, Zhou, and Chen}]{research}
Mingyang Chen, Linzhuang Sun, Tianpeng Li, Haoze Sun, Yijie Zhou, Chenzheng Zhu, Haofen Wang, Jeff~Z. Pan, Wen Zhang, Huajun Chen, Fan Yang, Zenan Zhou, and Weipeng Chen. 2025.
\newblock \href {https://arxiv.org/abs/2503.19470} {Research: Learning to reason with search for llms via reinforcement learning}.
\newblock \emph{Preprint}, arXiv:2503.19470.

\bibitem[{Edge et~al.(2025)Edge, Trinh, Cheng, Bradley, Chao, Mody, Truitt, Metropolitansky, Ness, and Larson}]{graphrag}
Darren Edge, Ha~Trinh, Newman Cheng, Joshua Bradley, Alex Chao, Apurva Mody, Steven Truitt, Dasha Metropolitansky, Robert~Osazuwa Ness, and Jonathan Larson. 2025.
\newblock \href {https://arxiv.org/abs/2404.16130} {From local to global: A graph rag approach to query-focused summarization}.
\newblock \emph{Preprint}, arXiv:2404.16130.

\bibitem[{Gao et~al.(2023)Gao, Yen, Yu, and Fried}]{gao2023verifiable}
Tianyu Gao, Howard Yen, Jiatong Yu, and Danqi Fried. 2023.
\newblock Enabling large language models to generate text with citations.
\newblock \emph{arXiv preprint arXiv:2305.14627}.

\bibitem[{Gutiérrez et~al.(2025)Gutiérrez, Shu, Qi, Zhou, and Su}]{hipporag}
Bernal~Jiménez Gutiérrez, Yiheng Shu, Weijian Qi, Sizhe Zhou, and Yu~Su. 2025.
\newblock \href {https://arxiv.org/abs/2502.14802} {From rag to memory: Non-parametric continual learning for large language models}.
\newblock \emph{Preprint}, arXiv:2502.14802.

\bibitem[{Ho et~al.(2020)Ho, Nguyen, Sugawara, and Aizawa}]{2wiki}
Xanh Ho, Anh-Khoa~Duong Nguyen, Saku Sugawara, and Akiko Aizawa. 2020.
\newblock \href {https://arxiv.org/abs/2011.01060} {Constructing a multi-hop qa dataset for comprehensive evaluation of reasoning steps}.
\newblock \emph{Preprint}, arXiv:2011.01060.

\bibitem[{Izacard et~al.(2022)Izacard, Caron, Hosseini, Riedel, Bojanowski, Joulin, and Grave}]{contriever}
Gautier Izacard, Mathilde Caron, Lucas Hosseini, Sebastian Riedel, Piotr Bojanowski, Armand Joulin, and Edouard Grave. 2022.
\newblock \href {https://arxiv.org/abs/2112.09118} {Unsupervised dense information retrieval with contrastive learning}.
\newblock \emph{Preprint}, arXiv:2112.09118.

\bibitem[{Jiang et~al.(2023)Jiang, Xu, Gao, Sun, Liu, Dwivedi-Yu, Yang, Callan, and Neubig}]{flare}
Zhengbao Jiang, Frank~F. Xu, Luyu Gao, Zhiqing Sun, Qian Liu, Jane Dwivedi-Yu, Yiming Yang, Jamie Callan, and Graham Neubig. 2023.
\newblock \href {https://arxiv.org/abs/2305.06983} {Active retrieval augmented generation}.
\newblock \emph{Preprint}, arXiv:2305.06983.

\bibitem[{Jin et~al.(2025)Jin, Zeng, Yue, Yoon, Arik, Wang, Zamani, and Han}]{search-r1}
Bowen Jin, Hansi Zeng, Zhenrui Yue, Jinsung Yoon, Sercan Arik, Dong Wang, Hamed Zamani, and Jiawei Han. 2025.
\newblock \href {https://arxiv.org/abs/2503.09516} {Search-r1: Training llms to reason and leverage search engines with reinforcement learning}.
\newblock \emph{Preprint}, arXiv:2503.09516.

\bibitem[{Joshi et~al.(2017)Joshi, Choi, Weld, and Zettlemoyer}]{triviaqa}
Mandar Joshi, Eunsol Choi, Daniel~S. Weld, and Luke Zettlemoyer. 2017.
\newblock \href {https://arxiv.org/abs/1705.03551} {Triviaqa: A large scale distantly supervised challenge dataset for reading comprehension}.
\newblock \emph{Preprint}, arXiv:1705.03551.

\bibitem[{Karpukhin et~al.(2020)Karpukhin, Oğuz, Min, Lewis, Wu, Edunov, Chen, and tau Yih}]{dpr}
Vladimir Karpukhin, Barlas Oğuz, Sewon Min, Patrick Lewis, Ledell Wu, Sergey Edunov, Danqi Chen, and Wen tau Yih. 2020.
\newblock \href {https://arxiv.org/abs/2004.04906} {Dense passage retrieval for open-domain question answering}.
\newblock \emph{Preprint}, arXiv:2004.04906.

\bibitem[{Kwiatkowski et~al.(2019)Kwiatkowski, Palomaki, Redfield, Collins, Parikh, Alberti, Epstein, Polosukhin, Devlin, Lee, Toutanova, Jones, Kelcey, Chang, Dai, Uszkoreit, Le, and Petrov}]{nq}
Tom Kwiatkowski, Jennimaria Palomaki, Olivia Redfield, Michael Collins, Ankur Parikh, Chris Alberti, Danielle Epstein, Illia Polosukhin, Jacob Devlin, Kenton Lee, Kristina Toutanova, Llion Jones, Matthew Kelcey, Ming-Wei Chang, Andrew~M. Dai, Jakob Uszkoreit, Quoc Le, and Slav Petrov. 2019.
\newblock \href {https://doi.org/10.1162/tacl_a_00276} {Natural questions: A benchmark for question answering research}.
\newblock \emph{Transactions of the Association for Computational Linguistics}, 7:452--466.

\bibitem[{Lewis et~al.(2021)Lewis, Perez, Piktus, Petroni, Karpukhin, Goyal, Küttler, Lewis, tau Yih, Rocktäschel, Riedel, and Kiela}]{rag}
Patrick Lewis, Ethan Perez, Aleksandra Piktus, Fabio Petroni, Vladimir Karpukhin, Naman Goyal, Heinrich Küttler, Mike Lewis, Wen tau Yih, Tim Rocktäschel, Sebastian Riedel, and Douwe Kiela. 2021.
\newblock \href {https://arxiv.org/abs/2005.11401} {Retrieval-augmented generation for knowledge-intensive nlp tasks}.
\newblock \emph{Preprint}, arXiv:2005.11401.

\bibitem[{Li et~al.(2025)Li, Luo, Li, Li, Cheng, Wang, Zheng, Wang, Yin, and Qiu}]{r3-rag}
Yuan Li, Qi~Luo, Xiaonan Li, Bufan Li, Qinyuan Cheng, Bo~Wang, Yining Zheng, Yuxin Wang, Zhangyue Yin, and Xipeng Qiu. 2025.
\newblock \href {https://arxiv.org/abs/2505.23794} {R3-rag: Learning step-by-step reasoning and retrieval for llms via reinforcement learning}.
\newblock \emph{Preprint}, arXiv:2505.23794.

\bibitem[{Luo et~al.(2025)Luo, E, Chen, Zheng, Wu, Guo, Lin, Feng, Kuang, Song, Zhu, and Tuan}]{hypergraphrag}
Haoran Luo, Haihong E, Guanting Chen, Yandan Zheng, Xiaobao Wu, Yikai Guo, Qika Lin, Yu~Feng, Zemin Kuang, Meina Song, Yifan Zhu, and Luu~Anh Tuan. 2025.
\newblock \href {https://arxiv.org/abs/2503.21322} {Hypergraphrag: Retrieval-augmented generation via hypergraph-structured knowledge representation}.
\newblock \emph{Preprint}, arXiv:2503.21322.

\bibitem[{Luo et~al.(2024)Luo, Liu, Xiao, Zhou, Chen, Zhao, and Liu}]{chunkfree}
Kun Luo, Zheng Liu, Shitao Xiao, Tong Zhou, Yubo Chen, Jun Zhao, and Kang Liu. 2024.
\newblock \href {https://doi.org/10.18653/v1/2024.acl-long.180} {Landmark embedding: A chunking-free embedding method for retrieval augmented long-context large language models}.
\newblock In \emph{Proceedings of the 62nd Annual Meeting of the Association for Computational Linguistics (Volume 1: Long Papers)}, pages 3268--3281, Bangkok, Thailand. Association for Computational Linguistics.

\bibitem[{Mallen et~al.(2023)Mallen, Asai, Zhong, Das, Khashabi, and Hajishirzi}]{popqa}
Alex Mallen, Akari Asai, Victor Zhong, Rajarshi Das, Daniel Khashabi, and Hannaneh Hajishirzi. 2023.
\newblock \href {https://arxiv.org/abs/2212.10511} {When not to trust language models: Investigating effectiveness of parametric and non-parametric memories}.
\newblock \emph{Preprint}, arXiv:2212.10511.

\bibitem[{OpenAI et~al.(2024)OpenAI, :, Jaech, Kalai, Lerer, Richardson, El-Kishky, Low, Helyar, Madry, Beutel, Carney, Iftimie, Karpenko, Passos, Neitz, Prokofiev, Wei, Tam, Bennett, Kumar, Saraiva, Vallone, Duberstein, Kondrich, Mishchenko, Applebaum, Jiang, Nair, Zoph, Ghorbani, Rossen, Sokolowsky, Barak, McGrew, Minaiev, Hao, Baker, Houghton, McKinzie, Eastman, Lugaresi, Bassin, Hudson, Li, de~Bourcy, Voss, Shen, Zhang, Koch, Orsinger, Hesse, Fischer, Chan, Roberts, Kappler, Levy, Selsam, Dohan, Farhi, Mely, Robinson, Tsipras, Li, Oprica, Freeman, Zhang, Wong, Proehl, Cheung, Mitchell, Wallace, Ritter, Mays, Wang, Such, Raso, Leoni, Tsimpourlas, Song, von Lohmann, Sulit, Salmon, Parascandolo, Chabot, Zhao, Brockman, Leclerc, Salman, Bao, Sheng, Andrin, Bagherinezhad, Ren, Lightman, Chung, Kivlichan, O'Connell, Osband, Gilaberte, Akkaya, Kostrikov, Sutskever, Kofman, Pachocki, Lennon, Wei, Harb, Twore, Feng, Yu, Weng, Tang, Yu, Candela, Palermo, Parish, Heidecke, Hallman, Rizzo, Gordon, Uesato, Ward,
  Huizinga, Wang, Chen, Xiao, Singhal, Nguyen, Cobbe, Shi, Wood, Rimbach, Gu-Lemberg, Liu, Lu, Stone, Yu, Ahmad, Yang, Liu, Maksin, Ho, Fedus, Weng, Li, McCallum, Held, Kuhn, Kondraciuk, Kaiser, Metz, Boyd, Trebacz, Joglekar, Chen, Tintor, Meyer, Jones, Kaufer, Schwarzer, Shah, Yatbaz, Guan, Xu, Yan, Glaese, Chen, Lampe, Malek, Wang, Fradin, McClay, Pavlov, Wang, Wang, Murati, Bavarian, Rohaninejad, McAleese, Chowdhury, Chowdhury, Ryder, Tezak, Brown, Nachum, Boiko, Murk, Watkins, Chao, Ashbourne, Izmailov, Zhokhov, Dias, Arora, Lin, Lopes, Gaon, Miyara, Leike, Hwang, Garg, Brown, James, Shu, Cheu, Greene, Jain, Altman, Toizer, Toyer, Miserendino, Agarwal, Hernandez, Baker, McKinney, Yan, Zhao, Hu, Santurkar, Chaudhuri, Zhang, Fu, Papay, Lin, Balaji, Sanjeev, Sidor, Broda, Clark, Wang, Gordon, Sanders, Patwardhan, Sottiaux, Degry, Dimson, Zheng, Garipov, Stasi, Bansal, Creech, Peterson, Eloundou, Qi, Kosaraju, Monaco, Pong, Fomenko, Zheng, Zhou, McCabe, Zaremba, Dubois, Lu, Chen, Cha, Bai, He, Zhang, Wang,
  Shao, and Li}]{openai2024openaio1card}
OpenAI, :, Aaron Jaech, Adam Kalai, Adam Lerer, Adam Richardson, Ahmed El-Kishky, Aiden Low, Alec Helyar, Aleksander Madry, Alex Beutel, Alex Carney, Alex Iftimie, Alex Karpenko, Alex~Tachard Passos, Alexander Neitz, Alexander Prokofiev, Alexander Wei, Allison Tam, and 244 others. 2024.
\newblock \href {https://arxiv.org/abs/2412.16720} {Openai o1 system card}.
\newblock \emph{Preprint}, arXiv:2412.16720.

\bibitem[{Pu et~al.(2024)Pu, He, Qiu, Wu, and Yu}]{noiserag}
Yuan Pu, Zhuolun He, Tairu Qiu, Haoyuan Wu, and Bei Yu. 2024.
\newblock \href {https://arxiv.org/abs/2407.15353} {Customized retrieval augmented generation and benchmarking for eda tool documentation qa}.
\newblock \emph{Preprint}, arXiv:2407.15353.

\bibitem[{Qwen et~al.(2025)Qwen, :, Yang, Yang, Zhang, Hui, Zheng, Yu, Li, Liu, Huang, Wei, Lin, Yang, Tu, Zhang, Yang, Yang, Zhou, Lin, Dang, Lu, Bao, Yang, Yu, Li, Xue, Zhang, Zhu, Men, Lin, Li, Tang, Xia, Ren, Ren, Fan, Su, Zhang, Wan, Liu, Cui, Zhang, and Qiu}]{qwen2.5}
Qwen, :, An~Yang, Baosong Yang, Beichen Zhang, Binyuan Hui, Bo~Zheng, Bowen Yu, Chengyuan Li, Dayiheng Liu, Fei Huang, Haoran Wei, Huan Lin, Jian Yang, Jianhong Tu, Jianwei Zhang, Jianxin Yang, Jiaxi Yang, Jingren Zhou, and 25 others. 2025.
\newblock \href {https://arxiv.org/abs/2412.15115} {Qwen2.5 technical report}.
\newblock \emph{Preprint}, arXiv:2412.15115.

\bibitem[{Rawte et~al.(2023)Rawte, Sheth, and Das}]{largefundataion}
Vipula Rawte, Amit Sheth, and Amitava Das. 2023.
\newblock \href {https://arxiv.org/abs/2309.05922} {A survey of hallucination in large foundation models}.
\newblock \emph{Preprint}, arXiv:2309.05922.

\bibitem[{Sanmartin(2024)}]{kg-rag}
Diego Sanmartin. 2024.
\newblock \href {https://arxiv.org/abs/2405.12035} {Kg-rag: Bridging the gap between knowledge and creativity}.
\newblock \emph{Preprint}, arXiv:2405.12035.

\bibitem[{Sarthi et~al.(2024)Sarthi, Abdullah, Tuli, Khanna, Goldie, and Manning}]{raptor}
Parth Sarthi, Salman Abdullah, Aditi Tuli, Shubh Khanna, Anna Goldie, and Christopher~D. Manning. 2024.
\newblock \href {https://arxiv.org/abs/2401.18059} {Raptor: Recursive abstractive processing for tree-organized retrieval}.
\newblock \emph{Preprint}, arXiv:2401.18059.

\bibitem[{Schwartz et~al.(2024)Schwartz, Choshen, Shtok, Doveh, Karlinsky, and Arbelle}]{numerologic}
Eli Schwartz, Leshem Choshen, Joseph Shtok, Sivan Doveh, Leonid Karlinsky, and Assaf Arbelle. 2024.
\newblock \href {https://doi.org/10.18653/v1/2024.emnlp-main.12} {{N}umero{L}ogic: Number encoding for enhanced {LLM}s' numerical reasoning}.
\newblock In \emph{Proceedings of the 2024 Conference on Empirical Methods in Natural Language Processing}, pages 206--212, Miami, Florida, USA. Association for Computational Linguistics.

\bibitem[{Trivedi et~al.(2022)Trivedi, Balasubramanian, Khot, and Sabharwal}]{musique}
Harsh Trivedi, Niranjan Balasubramanian, Tushar Khot, and Ashish Sabharwal. 2022.
\newblock \href {https://arxiv.org/abs/2108.00573} {Musique: Multihop questions via single-hop question composition}.
\newblock \emph{Preprint}, arXiv:2108.00573.

\bibitem[{Trivedi et~al.(2023)Trivedi, Balasubramanian, Khot, and Sabharwal}]{ircot}
Harsh Trivedi, Niranjan Balasubramanian, Tushar Khot, and Ashish Sabharwal. 2023.
\newblock \href {https://arxiv.org/abs/2212.10509} {Interleaving retrieval with chain-of-thought reasoning for knowledge-intensive multi-step questions}.
\newblock \emph{Preprint}, arXiv:2212.10509.

\bibitem[{Wang et~al.(2023{\natexlab{a}})}]{wang2023retrieval}
Alex Wang and 1 others. 2023{\natexlab{a}}.
\newblock When not to trust language models: Investigating effectiveness of parametric and non-parametric memories.
\newblock \emph{arXiv preprint arXiv:2212.10511}.

\bibitem[{Wang et~al.(2025)Wang, Tan, Dou, and Wen}]{omni}
Shuting Wang, Jiejun Tan, Zhicheng Dou, and Ji-Rong Wen. 2025.
\newblock \href {https://arxiv.org/abs/2412.13018} {Omnieval: An omnidirectional and automatic rag evaluation benchmark in financial domain}.
\newblock \emph{Preprint}, arXiv:2412.13018.

\bibitem[{Wang et~al.(2023{\natexlab{b}})Wang, Lipka, Rossi, Siu, Zhang, and Derr}]{multi-document}
Yu~Wang, Nedim Lipka, Ryan~A. Rossi, Alexa Siu, Ruiyi Zhang, and Tyler Derr. 2023{\natexlab{b}}.
\newblock \href {https://arxiv.org/abs/2308.11730} {Knowledge graph prompting for multi-document question answering}.
\newblock \emph{Preprint}, arXiv:2308.11730.

\bibitem[{Yang et~al.(2018)Yang, Qi, Zhang, Bengio, Cohen, Salakhutdinov, and Manning}]{hotpotqa}
Zhilin Yang, Peng Qi, Saizheng Zhang, Yoshua Bengio, William~W. Cohen, Ruslan Salakhutdinov, and Christopher~D. Manning. 2018.
\newblock \href {https://arxiv.org/abs/1809.09600} {Hotpotqa: A dataset for diverse, explainable multi-hop question answering}.
\newblock \emph{Preprint}, arXiv:1809.09600.

\bibitem[{Zhang et~al.(2025)Zhang, Li, Cui, Cai, Liu, Fu, Huang, Zhao, Zhang, Xu, Chen, Wang, Luu, Bi, Shi, and Shi}]{aiocean}
Yue Zhang, Yafu Li, Leyang Cui, Deng Cai, Lemao Liu, Tingchen Fu, Xinting Huang, Enbo Zhao, Yu~Zhang, Chen Xu, Yulong Chen, Longyue Wang, Anh~Tuan Luu, Wei Bi, Freda Shi, and Shuming Shi. 2025.
\newblock \href {https://arxiv.org/abs/2309.01219} {Siren's song in the ai ocean: A survey on hallucination in large language models}.
\newblock \emph{Preprint}, arXiv:2309.01219.

\end{thebibliography}

\end{document}